\pdfoutput=1

\documentclass[11pt]{article}

\usepackage{EMNLP2023}

\usepackage{times}
\usepackage{latexsym}

\usepackage[T1]{fontenc}

\usepackage[utf8]{inputenc}

\usepackage{microtype}

\usepackage{inconsolata}

\title{Cost-Effective Proxy Reward Model Construction with On-Policy and Active Learning }

\author{
Yifang Chen$^{1}$, Shuohang Wang$^{2}$, Ziyi Zhang$^{2}$, Hiteshi Sharma$^{2}$, \\
\bf Nikos Karampatziakis$^2$, Donghan Yu$^2$, Kevin Jamieson$^1$, Simon Shaolei Du$^1$, Yelong Shen$^2$ \\
$^{1}$University of Washington, Seattle \quad
$^{2}$Microsoft Corporation
}

\usepackage[utf8]{inputenc} 
\usepackage[T1]{fontenc}    
\usepackage{hyperref}       
\usepackage{url}            
\usepackage{booktabs}       
\usepackage{amsfonts}       
\usepackage{nicefrac}       
\usepackage{microtype}      
\usepackage{xcolor}         
\usepackage{color}
\usepackage{colortbl}
\usepackage{soul}
\usepackage{graphicx}
\usepackage{subfig}
\usepackage{amsmath}
\usepackage{amsthm}
\usepackage{xspace}
\usepackage{thmtools}
\usepackage{enumitem}
\usepackage{pifont}
\usepackage{multirow}
\usepackage{float}
\usepackage{titletoc}
\usepackage[titletoc]{appendix}
\usepackage{listings}
\usepackage{adjustbox}
\usepackage{mdframed}
\usepackage{natbib}

\usepackage{graphicx}
\usepackage{caption}
\usepackage{subcaption}

\definecolor{hl}{RGB}{205, 232, 248}

\definecolor{orange5}{RGB}{255, 235, 204}
\definecolor{orange20}{RGB}{255, 204, 153}
\definecolor{orange30}{RGB}{255, 190, 127}
\definecolor{orange40}{RGB}{255, 178, 102}
\definecolor{orange60}{RGB}{255, 153, 51}
\definecolor{orange80}{RGB}{255, 128, 10}
\definecolor{orange90}{RGB}{255, 115, 0}
\definecolor{orange100}{RGB}{255, 102, 0}

\usepackage{graphicx}
\usepackage{mathtools}
\usepackage{footnote}
\usepackage{float}
\usepackage{xspace}
\usepackage{multirow}
\usepackage{wrapfig}
\usepackage{framed}
\makesavenoteenv{tabular}
\makesavenoteenv{table}

\usepackage{amsmath}
\usepackage{amsfonts}
\usepackage{amssymb}
\usepackage{amsthm}
\usepackage{bm}
\usepackage{bbm}
\usepackage{mathtools}
\usepackage{enumitem}
\usepackage{thmtools,thm-restate}
\usepackage{algorithm}
\usepackage{algorithmic}
\usepackage{graphicx}
\usepackage{comment}

\def\CC{\mathcal{C}}

\def\RR{\mathcal{R}}

\def\*{\star}

\usepackage{graphicx}
\usepackage{mathtools}
\usepackage{footnote}
\usepackage{float}
\usepackage{xspace}
\usepackage{multirow}
\usepackage{wrapfig}
\usepackage{framed}
\usepackage{xcolor}

\newcommand{\IFTseed}{\text{IFT}_\text{seed}}
\newcommand{\EFTseed}{\text{EFT}_\text{seed}}
\newcommand{\meval}{M^\text{eval}}

\usepackage{float}

\begin{document}
\maketitle
\begin{abstract}

Reinforcement learning with human feedback (RLHF), as a widely adopted approach in current large language model pipelines, is \textit{bottlenecked by the size of human preference data}. While traditional methods rely on offline preference dataset constructions, recent approaches have shifted towards online settings, where a learner uses a small amount of labeled seed data and a large pool of unlabeled prompts to iteratively construct new preference data through self-generated responses and high-quality reward/preference feedback. However, most current online algorithms still focus on preference labeling during policy model updating with given feedback oracles, which incurs significant expert query costs.
\textit{We are the first to explore cost-effective proxy reward oracles construction strategies for further labeling preferences or rewards with extremely limited labeled data and expert query budgets}. Our approach introduces two key innovations: (1) on-policy query to avoid OOD and imbalance issues in seed data, and (2) active learning to select the most informative data for preference queries. Using these methods, we train a evaluation model with minimal expert-labeled data, which then effectively labels nine times more preference pairs for further RLHF training. For instance, our model using Direct Preference Optimization (DPO) gains around over 1\% average improvement on AlpacaEval2, MMLU-5shot and MMLU-0shot, with only 1.7K query cost. Our methodology is orthogonal to other direct expert query-based strategies and therefore might be integrated with them to further reduce query costs.
%
%
\end{abstract}
\section{Introduction}
Reinforcement learning from human feedback (RLHF) has gained significant attention in recent years. Traditional approaches represented by Proximal Policy Optimization (PPO) \cite{ouyang2022training}, maintains one or several standalone reward models to finetune the policy model online by maximizing the rewards. Recently, people start to using Direct Preference Optimization (DPO) \cite{rafailov2024direct} and its variants due to their stable training properties. Some approaches query preferences directly from experts (e.g., humans, GPT) while others utilize a cheaper, offline-trained reward/preference model as a proxy oracle. However, all these methods suffer from the \textit{scarcity of human preference-labeled data.}

\begin{figure*}[!h]
    \centering
    \setlength{\belowcaptionskip}{-18pt} 
    \includegraphics[scale=0.52]{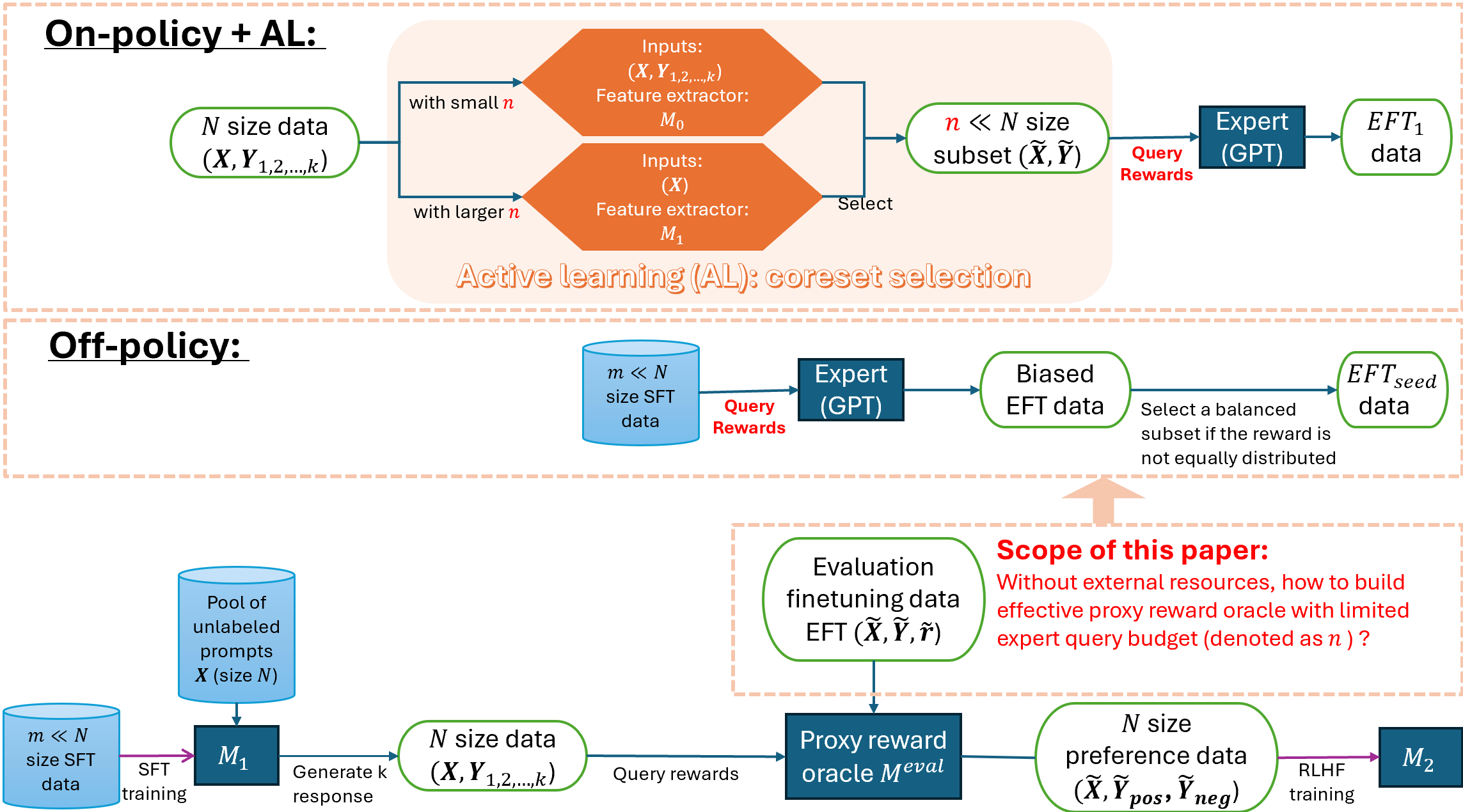}
    \caption{\textbf{Our cost-effective proxy reward oracle construction pipeline:} \textbf{Our main approach is shown as On-policy+AL} that features two innovations: an on-policy query framework that uses $M_1$ generated data to query preferences and train the evaluation model $\meval$, and (2) An active learning (AL) module that further aids in selecting $n \ll N$ budget informative data points. We also test\textbf{ Off-policy method}, which is adapted from self-rewarding LM \cite{yuan2024selfreward}. Unlike our on-policy query method, this approach queries the expert with seed SFT data and generally outperformed by \textbf{On-policy+AL} unless in the benign conditions. Note that our experiments build upon DPO training but this proxy oracle itself independent of the RLHF training method.}
    \label{fig:pipeline}
\end{figure*}

Classic works such as \citet{bai2022training, cui2023ultrafeedback, zhu2023starling} aim to build high-quality, model-independent preference datasets offline. However, these methodologies can lead to distribution shift issues when the training model differs from the exploratory models used to generate the dataset. Recent research has shifted focus to online techniques, also referred to as "self-improvement" or "iterative" methods \cite{wang2023self, yuan2024selfreward, rosset2024direct, xiong2023iterative, dong2024rlhf, viethoangtranduong, wu2024SPO, xu2023some, xie2024exploratory, chen2024bootstrapping}. These methods leverage a set amount of labeled seed data and a large pool of unlabeled prompts, with the goal of continuously constructing new preference data through responses generated by the model itself, and potentially through external reward/preference feedback. The primary cost in this process comes from feedback queries from experts.

Despite advances, most current online methods still focus on saving expert query costs directly for policy model training with fixed preference feedback oracles, as described in Sec.\ref{sec: existing works and baselines} and App.\ref{app: more realted works}. Given the high complexity of generative models, they either demand significant amounts of preference/reward labeling from expensive experts or rely on offline-trained reward models like PairRM \cite{jiang2023llm}, which itself requires substantial high-quality preference data. Conversely, the cost-saving strategies for  \textit{constructing the proxy reward oracle remain under-explored.}

Inspired by the success of proxy reward oracles, we hypothesize that a smaller dataset is sufficient to train a weak evaluation model that can effectively label a much larger set of RLHF data. Furthermore, inspired by the successes of online methods, we incorporate on-policy query techniques into the training of evaluation models. The term "on-policy," although not equivalent, is a key part of the "online" pipeline, as it emphasizes constructing the data using the target model being trained.

In this paper, we focus on cost-effective labeling strategies through constructing proxy reward oracles with only a limited amount of labeled seed data and a small expert query budget. For instance, our approach uses seed SFT data that is more than 10 times smaller than many existing works, and the query budget is on the same order as the seed data. Our objective is not to develop a new state-of-the-art model but to propose a methodology under these stringent conditions that can potentially be combined with other methods. The most closely related work to our study is the self-rewarding LM \cite{yuan2024selfreward}, where a cost-efficient method is used in training a proxy reward oracle but with a different focus and setting than ours. We also investigate a modified off-policy version by adapting their methods to our setting.

We highlight our proposed pipelines in Fig.\ref{fig:pipeline}. 
Specifically, our contributions can be summarized as threefold:

\begin{itemize}[leftmargin=*, itemsep=-2pt, before=\vspace{-8pt}, after=\vspace{-6pt}]
    \item We first propose a \textbf{random on-policy} expert query strategy for labeling the preference data used to train the proxy reward oracle. Our empirical study validates that a weak evaluation model trained with a small amount of data can effectively label about \textit{nine times} more preference pairs. For instance, with only \textit{1.7K query budget}, DPO training on Llama-2 7B with 15K preference pairs labeled by us yields over a 1\% increase in performance on AlpacaEval2 and MMLU 5-shot metrics compared to the initial supervised fine-tuning model. In comparison, directly using the queried rewards to label the preference data without training an proxy oracle result in less than a 0.1\% improvement under the same query budget. (Fig.~\ref{fig:unlabeledprompt_balance})
    \item Building on the success of the on-policy approach, we further explore replacing the random query strategy with \textbf{coreset-type} (e.g., \(k\)-center) active learning strategies, to select the most informative prompts and responses from the large unlabeled. Our active learning strategy results in additional performance gains from \textbf{random on-policy} strategy of 0.34\% to 0.6\% on the MMLU-5shot and 0.4\% to 0.53\% on MMLU-shot metrics under a properly managed budget. (Tab.~\ref{tab:coreset})
    \item Lastly, we also investigate other methods, such as \textbf{off-policy data query} strategy derived from the self-rewarding LM \cite{yuan2024selfreward} and variants of Self-play finetuning (\textbf{SPIN}) \cite{wang2023self} . Note that they are not directly comparable to our setting but help supporting the advantage of our on-policy + AL design, (Tab.~\ref{tab:coreset}, Sec.~\ref{sec: ablation more comparison with previous})
\end{itemize}

\section{Related works}
\label{sec: existing works and baselines}

\paragraph{Training without reward oracle.} The Self-play finetuning (SPIN) \cite{chen2024self} relies solely on seed SFT data by consistently pairing a ground truth response as a positive sample with a model-generated response as a negative, thereby obviating the need for expert queries. However, the efficacy of this method heavily depends on the availability of a large seed data volume to avoid over-fitting. For example, while the original SPIN methodology utilizes a dataset of 50K IFT instances, our study is constrained to a considerably smaller set of only 3K instances.  The pipeline is shown in Fig.\ref{fig: pipeline_previous}.

\paragraph{Using seed SFT data to train proxy oracle}
\citet{yuan2024selfreward} proposed a method for training the evaluation model using SFT seed data within a self-rewarding framework. Although their experimental setup differs from ours, their strategies can be adapted to our context as \textbf{off-policy query} detailed in Sec.~\ref{sec: baseline}. Specifically, they use a single model to serve both as the policy and evaluation models, generating evaluations of seed data from the initial SFT model and then updating the model based on self-generated data. However, the success of this self-iterative process relies on a significantly more powerful model, LLama2-70B, whereas our focus is a more general methodology for any model, including weaker models like Llama2-7B. To adjust for this disparity, we query GPT for evaluating the seed data and use the generated evaluation-inclusive data to train a standalone evaluation model. Another adaptation is that the original paper uses Llama2-70B-chat to generate instructions relevant to the seed data to avoid distribution shift. However, this should be counted into the expert query budget. Here, we replace this self-instruct step with fixed pool of unlabeled prompts in our setting. The pipeline is shown in Fig.\ref{fig:pipeline}.

\paragraph{Using external resources to train proxy oracle}
Directly querying preference or reward feedback from experts during the online process is expensive. Existing works like \cite{wu2024SPO, viethoangtranduong} utilized an offline-trained model, PairRM, proposed by \citet{jiang2023llm} as a proxy feedback oracle. Recently, \citet{dong2024rlhf} further trained three types of reward models: llm-as-judge, preference model, and Bradley-Terry based reward model, using a large mixture of offline datasets, and then selected the proper one using RewardBench \cite{lambert2024rewardbench}. We will NOT COMPARE with these methods as they uses external resources.

Many other methods focus on efficient query strategies for policy model training directly, with fixed reward/preference oracle. We postpone the details into App.~\ref{app: more realted works}.

\section{Proxy-reward-oracle based self-improvement with limited data}

Given a pretrained model \(M_0\), our approach assumes a small set of labeled seed SFT data in the format (instruction, response, reward). Note that the reward label is optional, since most standard instruction fine-tuning datasets do not contain reward information. In such cases, using rewards for seed data will also require an expert query budget. Additionally, we have access to a large pool of unlabeled instruction-only data, \(\bm{X}\), and expert which provides preference feedback (e.g., GPT-4). Note that \(\bm{X}\) is sourced differently from the seed data and therefore has a different distribution.

Our goal is to label \(\bm{X}\) by efficiently leveraging the expert's feedback and the intrinsic capabilities of \(M_0\). In practice, it is not always feasible to label $\bm{X}$ by querying superior LLMs such as GPT, considering the cost to label (large-scale) data can be formidable. Therefore, we propose to efficiently build a reward feedback oracle \(M^\text{eval}\) as a proxy to assist in labeling \(\bm{X}\), while minimizing the expert querying cost as much as possible. The performance of this proxy oracle will be measured by the final performance of the model trained on the newly labeled preference data. Since the problem setting is with strictly constrained budget to query expert LLMs, then using external datasets and benchmarks to train the proxy reward oracle (e.g., related works mentioned in Sec~\ref{sec: existing works and baselines}) is out of the scope.


Essentially, we utilize two types of data during the entire process. Following the same notation as \citet{yuan2024selfreward}, we use 
Instruction Fine Tuning (IFT)
to denote samples with the format \textit{[prompt, response]} for policy model training that generates instruction-following responses. On the other hand, we use 
Evaluation Fine Tuning (EFT) to denote samples with the format \textit{[prompt + response + evaluation criterion, justification + reward score (0-5)]} to train an evaluation model (i.e., the proxy reward oracle) that provides reward-inclusive evaluations for any given IFT pair. A detailed example of EFT is shown in Appendix~\ref{app: example of EFT}. Note that, unlike many existing works whose reward oracle yields numerical feedback only, we adopt the llm-as-judge framework \cite{zheng2024judging}, where the evaluation model itself is also a text generator.

\section{Our strategy: active on-policy query.}
\label{sec: strategy}
Now we are ready to present our on-policy active EFT query strategies, starting with the detailed pipelines as follows. (See visualization in Fig.~\ref{fig:pipeline}.)
\vspace{4px}
\paragraph{Detailed steps of our on-policy +AL pipeline. }
\begin{enumerate}[leftmargin=*, itemsep=-2pt, before=\vspace{-4pt}, after=\vspace{-3pt}]
\item Given the pretrained model $M_0$ and the initial seed data $\text{IFT}_\text{seed}$, SFT on $\text{IFT}_\text{seed}$ (or only on its high reward part if available) to obtain $M_1$. 
\item Given a set of $N$ unlabeled prompts $\mathbf{X}$, for \textbf{each} $x \in \mathbf{X}$, generate a set of $k$ responses $\tilde{y}_0, \tilde{y}_1, \ldots, \tilde{y}_k$ using $M_1$. Denote the entire pool of responses as $\tilde{\mathbf{Y}}_1$ and the whole $N*k$ size generated samples as $\text{IFT}_1$.
\vspace{4px}
\item 
\begin{mdframed}[backgroundcolor=blue!10, linewidth=0.5pt, linecolor=blue, innertopmargin=3pt, innerbottommargin=0pt, innerrightmargin=3pt, innerleftmargin=3pt]
Use active query strategies (explained below) to select a $n \ll N*k$ budget subset of $\text{IFT}_1$, query expert (e.g. GPT) for their evaluation results based on the evaluation criterion templates, and therefore construct $\text{EFT}_1$.
\end{mdframed}
\item Based on a pretrained model $M_0$, SFT on $\text{EFT}_1$ to get a weak evaluation model $M_1^{\text{eval}}$.
\item Generate rewards for the rest of unqueried $\text{IFT}_1$ using $M_1^{\text{eval}}$. For each prompt, choose the highest and lowest samples to form a DPO pair and denote the whole set as $\text{DPO}_1$
\item Finally trained $M_2$ based on $M_1$ using $\text{DPO}_1$.
\end{enumerate}
The key contribution of our pipeline comes from the third step. Firstly, we emphasize \textbf{on-policy} EFT querying, where the term 'on-policy' refers to sampling from the target model we are training, rather than utilizing external resources. I.e., we generate $\text{EFT}_1$ based on responses from the policy model $M_1$, rather than relying on the initial $\text{EFT}_\text{seed}$ . Secondly, rather than randomly selecting a subset of $\text{IFT}_1$ for querying, we employ \textbf{Active Learning (AL) strategies} to identify and select a more informative subset.

\noindent\textbf{Focus on one iteration and DPO}
In this study, we limit our focus to a single iteration, rather than multiple iterations, to specifically analyze the impact of on-policy querying and AL strategies. It is entirely feasible to extend our pipeline to multiple iterations, the single-iteration here allows us to isolate and understand the effects more clearly. 

\subsection{Random (passive) on-policy query}
The previous method involves training $\tilde{M}^\text{eval}$ via $\text{EFT}_\text{seed}$. However, this approach faces two main challenges: Firstly, due to the distribution shift from seed data to unlabeled prompts $X$, the $\text{IFT}_1$ generated by the policy model may fall into the out-of-distribution (OOD) domain of evaluation model. Secondly, we observe that the reward distribution for seed data is often biased towards higher values. This bias arises because $\text{EFT}_\text{seed}$, derived from $\text{IFT}_\text{seed}$, typically consists of human-annotated, high-quality entries, which benefits the SFT phase but can leads to over-fitting when training evaluation models. An example of this can be seen in the left part of Figure~\ref{fig:reward_distr} (specific dataset details will be provided later). Training with a balanced reward distribution can mitigate such bias issues, but also significantly reduces the effective number of training EFTs (e.g., less than 20\% of total $\text{EFT}_\text{seed}$ are used for training $M^\text{eval}$), thus limiting potential improvements.

To address these two problems, we first propose \textbf{random on-policy query} for constructing $M^\text{eval}$. This method involves randomly selecting a subset of $n$ prompts from $\bm{X}$ and generating responses using our target policy $M_1$ (i.e. on-policy). This not only avoid OOD problem when using $M^\text{eval}$ to label the rest of $\text{IFT}_1$, but also natural leads to more balanced rewards distribution among $\text{EFT}_1$ as shown in the right part of Fig.~\ref{fig:reward_distr}. 

\paragraph{Variant: random on-policy query with balanced training.} Although $\text{EFT}_1$ exhibits a more diverse reward distribution than $\text{EFT}_\text{seed}$, we can further enforce the strict balance by setting the number of samples for each reward score to be equal. Later we show that the unbalanced and balanced version each may leads to different advantages.

\subsection{Active on-policy query: coresetEFT and coresetIFT}
Active querying for LLM finetuning has been studied in \citet{bhatt2024experimental} but they focused on query response for IFT dataset that is for supervised finetuning. Their results show that actively learning a generative model is challenging. However, our goal here is to actively learn a weak evaluator $M_1^\text{eval}$, which is more close to classification tasks, where numerous of AL strategies has been proved to be effective.(e.g. 
\citet{k_center_coreset,geifman2017deep,citovsky2021batch,ash2019deep,ash2021gone}) 
The similar conjecture has also been proposed in \citet{dong2024rlhf} where they believe that, reward feedback, as a discriminative task, might be easier than generative tasks

While there exists many AL strategies, here we focused on the classical coreset selection. (In some place, people will call this K-center selection.) The main idea is to annotate inputs that are \emph{diverse} in the representation space.
\citet{k_center_coreset} proposed a k-center objective that chooses $k$ size subset $S$ as centers of balls with equal radius:
\vspace{-8px}
\begin{equation}
\vspace{-4px}
\label{eq:k-center}
    S  = \operatorname*{argmin}_{ S' \subset X,|S'| = k}    \operatorname*{max}_{\substack{i \in X}} \operatorname*{min}_{\substack{j \in S'}} \lVert f(x_i)-f(x_j)\rVert,
\end{equation}
where $f$ is a feature extractor that maps prompts into feature space in $\mathbb{R}^d$ and is derived from the pretrained model $h$. For decoder-only architectures, we use the first hidden state as the feature. To optimize this NP-hard objective \citep{cook1998combinatorial}, we follow the greedy methods proposed by \citet{k_center_coreset}, which enjoy a 2-multiplicative approximation guarantee to the optimal selection.

\paragraph{Two ways of extracting embedding.}
In classical AL problems, the inputs and embedding models are straightforward since the queries are trained on the same data as the model whose final performance is of interest, and the training set inputs are fixed. In our scenario, we are not directly comparing the performance of $M_1^\text{eval}$, and part of $\text{EFT}_1$ is generated by the model itself. Therefore, we propose two methods for extracting embeddings and study the benefits of each.
\begin{itemize}[leftmargin=*,itemsep=-2pt, before=\vspace{-7pt}, after=\vspace{-7pt}]
    \item \textbf{coresetEFT:} We use the instruction for each EFT sample (i.e IFT prompts + $M_1$ generated response + evaluation criterion).
    \item\textbf{coresetIFT:}  We use the instruction for each IFT sample as input and used seed IFT trained $M_1$ as the embedding extractor model. 
\end{itemize}
While both methods involves information provided by unlabeled prompts and the SFT trained $M_1$, the coresetEFT explicitly consider the embedding of the generated outputs. Therefore, coresetEFT can be reduced to standard active learning for discriminative problem (i.e classification) where the learner take aims to find the decision boundary a. On the other hand, the second coresetIFT makes assumption that, the evaluation made by trained evaluator mainly depends on the prompts instead of the generated response.

\subsection{Summary of our approaches}
As summary, we proposed three approaches -- \textbf{random on-policy} and two active on-policy strategies \textbf{coresetIFT} and \textbf{coresetEFT}. We also adapt the SPIN and self-rewarding methods mentioned in Sec.~\ref{sec: existing works and baselines} to our settings as additional investigation. We will focus on the following three questions:
\begin{itemize}[leftmargin=*, itemsep=-3pt, before=\vspace{-8pt}, after=\vspace{-4pt}]
\item \textbf{Q1.} Is a weak evaluator $M^\text{eval}$ trained on a small budget $n$ of $\text{EFT}_1$ sufficient to construct a larger preference set, and is the performance of $M_2$ always positively correlated with the size $n$?
\item \textbf{Q2.} Can an active learning strategy further improve the performance over random on-policy?
\item \textbf{Q3.} How does on-policy+AL strategy compare with other candidate approaches like off-policy query and variants of SPIN?
\end{itemize}

\section{Experiments}

\subsection{Experimental setup}
\label{sec: general experiment setting}

\paragraph{Models and dataset} We choose pretrained model $M_0$ to be Llama-2 7B \cite{touvron2023llama} and the first round conversion of OpenAssistant (oasst1) \cite{kopf2024openassistant} as the initial $\text{IFT}_\text{seed}$ whose size is around 3K. We specifically select the high-reward data from oasst1 as SFT data. For $\text{EFT}_\text{seed}$, the original reward scores from oasst1 lacked justification and did not conform to our evaluation templates. Consequently, we constructed an $\text{EFT}_\text{seed}$ by querying GPT using all 3K $\text{IFT}_\text{seed}$. (Only when applying the off-policy query approach.)
For unlabeled prompts $\bm{X}$,
we selected random $N$ subsets of prompts, ranging from 2.8K to 16.8K, from the Supernatural Instruction dataset \cite{wang2022super} and generate $k=4$ responses for each prompt.


\paragraph{Train weak evaluator $M_1^\text{eval}$} For each EFT dataset used to train $M_1^\text{eval}$, we randomly selected 300 (or 200 when train with $\text{EFT}_\text{seed}$) sample as validation set, with the remainder forming the training set to address training randomness. Specifically, we trained $M_1^\text{eval}$ using EFT train set over three random seeds for three epochs, and choose the best checkpoint using the validation set. 

\paragraph{Randomness and the impact of initial SFT model}
We trained three different versions of $M_1$ using the same $\text{IFT}_\text{seed}$ but with varying random seeds to mitigate training randomness and to explore the influence of the initial model quality on the data synthesis pipeline. Each $M_1$ version was then used to generate responses and construct DPO pairs. Each DPO set was subsequently trained with three random seeds. In all the results in the rest of the paper, unless specified, we report the average accuracy and sometimes square root of the total variance (denote as $\sqrt{tv}$) across all nine random seeds.
$
    tv = \mathbb{E}[\text{Var}(M_2 \mid M_1)] + \text{Var}(\mathbb{E}[M_2 \mid M_1])
$

\vspace{6px}
\noindent\textbf{Evaluation metric}~~
We evaluate the performance of our EFT query strategy by measuring the performance change from the initial SFT model $M_1$ to the final policy model $M_2$. Here we first use AlpacaEval2 \citet{alpaca_eval}, MMLU-0shot, MMLU-5shot \cite{suzgun2023challenging} as the downstream metrics to assess the performance of three proposed strategies. Then, we add BBH-COT and BBH-NOCOT \citet{hendrycks2020measuring}  where the prompts from supernatural instruction is less relevant to further investigate those methods through ablation studies.

We postpone more details in Appendix~\ref{app: experiment setup}.


\subsubsection{Other approaches investigated in ablation studies}
\label{sec: baseline}
We compare our approach with SPIN and off-policy query as explained in Sec.~\ref{sec: existing works and baselines}. Both methods have different original settings from ours, so we adapt and reproduce their approaches in our setting.

\paragraph{Train with $\text{EFT}_\text{seed}$ (Off-policy query)}
Due to the high bias in $\text{EFT}_\text{seed}$, we select only 200 samples among the all 3K queried $\EFTseed$ as the training set to ensure an equal number of rewards per class during training. The number of query budgets and exact query strategies under this setting depends on whether the initial rewards of $\text{IFT}_\text{seed}$ are known or not. Suppose the rewards of $\IFTseed$ are roughly known in advance; then we only need to query and construct $\EFTseed$ for an equal number of samples for each reward class, leading to a 200 (train) + 300 (validation) query budget. We refer to this method as \textbf{balanced off-policy query}. Otherwise, if the rewards are unknown, which is common in most SFT datasets, then we need to query the entire set of seed SFT data to find 200 balanced samples, given that $\IFTseed$ is highly biased. We refer to this method as \textbf{off-policy query}.

\paragraph{SPIN and its variants}
We not only compare the original SPIN with our proposed methods, but also highlight the disadvantages of SPIN \textit{under the setting where no unlabeled prompts are available}, as shown in Tab.~\ref{tab: performance comparison with SPIN}. Specifically, we train $\widetilde{M}^\text{eval}$ using $\text{EFT}_\text{seed}$ and employ it to evaluate responses generated by \(M_1\) for each prompt in $\text{IFT}_\text{seed}$ instead of $\bm{X}$. 

For the original \textbf{SPIN}, we choose human-annotated responses from oasst1 (i.e., $\text{IFT}_\text{seed}$) as positive samples and randomly generated $\widetilde{y}$ by $M_1$ as negative samples. For a hybrid version, denoted as \textbf{SPIN+$\bm{\widetilde{M}^\text{eval}}$}, we use the ground truth response as the positive sample and the $\widetilde{M}^\text{eval}$ generated response with the lowest reward as the negative one. Finally, for the method denoted as $\bm{\widetilde{M}^\text{eval}}$, both positive and negative samples are selected based on their evaluation by \(\widetilde{M}^\text{eval}\).


\subsection{Main Results: performance across different strategy and query budget $n$}
\label{sec: main results}

\begin{table}[!ht]
\centering
\setlength{\belowcaptionskip}{-2pt} 
\scalebox{0.56}{
\begin{tabular}{lccc}
\toprule
 & AlpacaEval2 & MMLU5shot & MMLU0shot \\
\midrule
Performance of $M_1$ & 3.15 ($\sqrt{tv}$ 0.02) & 43.11 ($\sqrt{tv}$ 1.6) & 42.46 ($\sqrt{tv}$ 0.79) \\
\bottomrule
\end{tabular}
}
\scalebox{0.6}{
\begin{tabular}{lp{2.6cm}p{2.6cm}p{2.6cm}}
\toprule
 & \textbf{random on-policy}(ours)  & \textbf{coresetEFT}  (ours) & \textbf{coresetIFT}  (ours) \\
\midrule
\midrule
\multicolumn{4}{c}{query budget n = 400} \\
\midrule
AlpacaEval2  & \cellcolor{orange!30} +0.61($\sqrt{tv}$0.53) & \cellcolor{orange!20} +0.54($\sqrt{tv}$0.62) & \cellcolor{orange!40} +0.7($\sqrt{tv}$0.40) \\
MMLU5shot   & \cellcolor{blue!20}-0.83($\sqrt{tv}$0.39) & -0.04($\sqrt{tv}$0.36) & \cellcolor{blue!20}-0.9($\sqrt{tv}$0.62) \\
MMLU0shot   & \cellcolor{orange!20} +0.19($\sqrt{tv}$0.34) & \cellcolor{orange!40} +0.31($\sqrt{tv}$0.23) & +0.02($\sqrt{tv}$0.19) \\
\midrule
\multicolumn{4}{c}{query budget n=1200} \\
\midrule
AlpacaEval2   & \cellcolor{orange!100} +1.42($\sqrt{tv}$0.50) & \cellcolor{orange!80} +0.99($\sqrt{tv}$0.47) & \cellcolor{orange!60} +0.85($\sqrt{tv}$0.56) \\
MMLU5shot   & \cellcolor{orange!60} +0.62($\sqrt{tv}$0.56) & \cellcolor{orange!90} +1.02($\sqrt{tv}$0.77) & \cellcolor{orange!80} +0.79($\sqrt{tv}$0.46) \\
MMLU0shot   & \cellcolor{orange!40} +0.35($\sqrt{tv}$0.76) & \cellcolor{orange!100} +0.88($\sqrt{tv}$0.55) & \cellcolor{orange!40} +0.35($\sqrt{tv}$0.29) \\
\midrule
\multicolumn{4}{c}{query budget n=1700} \\
\midrule
AlpacaEval2   & \cellcolor{orange!90}+1.18($\sqrt{tv}$0.57)& \cellcolor{orange!90}+1.24($\sqrt{tv}$0.62)& \cellcolor{orange!90}+1.33($\sqrt{tv}$0.56) \\
MMLU5shot   & \cellcolor{orange!90}+1.00($\sqrt{tv}$0.25)& \cellcolor{orange!100}+1.38($\sqrt{tv}$0.26) & \cellcolor{orange!100}+1.26($\sqrt{tv}$0.55) \\
MMLU0shot & \cellcolor{orange!20}+0.10($\sqrt{tv}$0.81) & -0.07($\sqrt{tv}$1.58) & \cellcolor{orange!60}+0.54($\sqrt{tv}$0.53) \\
\midrule
\multicolumn{4}{c}{query budget n=4800 } \\
\midrule
AlpacaEval2 & \cellcolor{orange!60} +0.86($\sqrt{tv}$0.77)& \cellcolor{orange!40} +0.73($\sqrt{tv}$0.76)& \cellcolor{orange!60}+0.82($\sqrt{tv}$0.5) \\
MMLU5shot & \cellcolor{orange!90} +1.2($\sqrt{tv}$0.60)& \cellcolor{orange!100}+1.54($\sqrt{tv}$0.51)& \cellcolor{orange!100}+1.54($\sqrt{tv}$0.14) \\
MMLU0shot & -0.23($\sqrt{tv}$1.73) & \cellcolor{orange!20}+0.22($\sqrt{tv}$1.08) & \cellcolor{orange!60} +0.45($\sqrt{tv}$0.45) \\
\bottomrule
\end{tabular}
}
\caption{\textbf{With fixed $N$, performance change from $M_1$ to $\widetilde{M}_2$ among our different strategies.} This table presents comprehensive results for three proposed methods across four different query budgets. It is easy to see, while random on-policy already gives positive result, the two active learning strategies gives further improvements.}
\label{tab:coreset}
\end{table}

With a fixed number of unlabeled prompts at N=16.8K, we evaluate the performance of \textbf{random on-policy}, \textbf{coresetIFT on-policy}, and \textbf{coresetEFT on-policy} strategies across various 
$n$ budgets using the AlpacaEval2 and MMLU metrics. As shown in Tab.~\ref{tab:coreset}, our random on-policy strategies generally result an effective proxy reward oracle, and AL module further improves performance. This answers \textbf{Q2}. More comparisons with other candidate approaches are investigated in Sec.~\ref{sec: ablation more comparison with previous}. Below, we provide further discussion on the performance of our approaches.

\paragraph{Over-Optimizing \(M_1^\text{eval}\) Can Degrade Performance.}
For all three methods, a consistent increase is observed only in the MMLU5shot metric. In contrast, for both AlpacaEval2 and MMLU0shot, we observe that an initial increase performance begins to decline after reaching a budget of 1000 or 1500, despite the increasing validation accuracy on \(M_1^\text{eval}\), therefore partially answering the \textbf{Q1} that the performance of $M_2$ is not always positively correlated with $n$. We believe this phenomenon is similar to what has been observed by \cite{moskovitz2023confronting}, where they show that with a fixed reward model, accumulating higher rewards past a certain point is associated with worse performance. Here, we are not directly maximizing the reward but using a proxy reward oracle to construct DPO pairs. We believe that noise in weak $M^\text{eval}$ implicitly serves as a regularization to avoid over-optimization. We will further investigate the correlation between $M^\text{eval}$ and other metrics in Sec.~\ref{sec: ablation study}.

\paragraph{Use coresetEFT at low budget and coreserIFT otherwise.}
On AlpacaEval2, the random strategy gains a slight advantage at lower budgets, but overall, all strategies perform similarly when considering the large standard deviation from AlpacaEval2. Conversely, both coreset strategies exhibit larger improvements than the random strategy on MMLU5shot and MMLU0shot. 
Now when comparing two embedding methods, we see that CoresetEFT demonstrates a dominant advantage at budgets of 200 and 1000, but these advantages diminish as the budget increases. In contrast, CoresetIFT, despite initially lower performance compared to the other two strategies, exhibits steady improvements across all metrics as the budget increases, eventually outperforming CoresetEFT. Notably, it achieves significantly lower total variance on MMLU0shot compared to the other methods.


\subsection{Ablation study: Sufficiency of training weak evaluator with low budget EFT} 
We have demonstrated the advantages of on-policy and active learning strategies with a fixed $N=16.8K$, using the AlpacaEval2 and MMLU metrics. Here, we further study the labeling ability of $\meval$ across different values of $N$ with a fixed query budget $n$ to address \textbf{Q1}, which give an affirmative answer that with limited budget $n$, using that to train a proxy oracle is better than direct label preference. In this ablation study, we focus on the random on-policy strategy as it generally exhibits similar trends to AL.

\begin{figure*}[!h]
    \centering
    \setlength{\belowcaptionskip}{-14pt} 
    \includegraphics[scale=0.35]{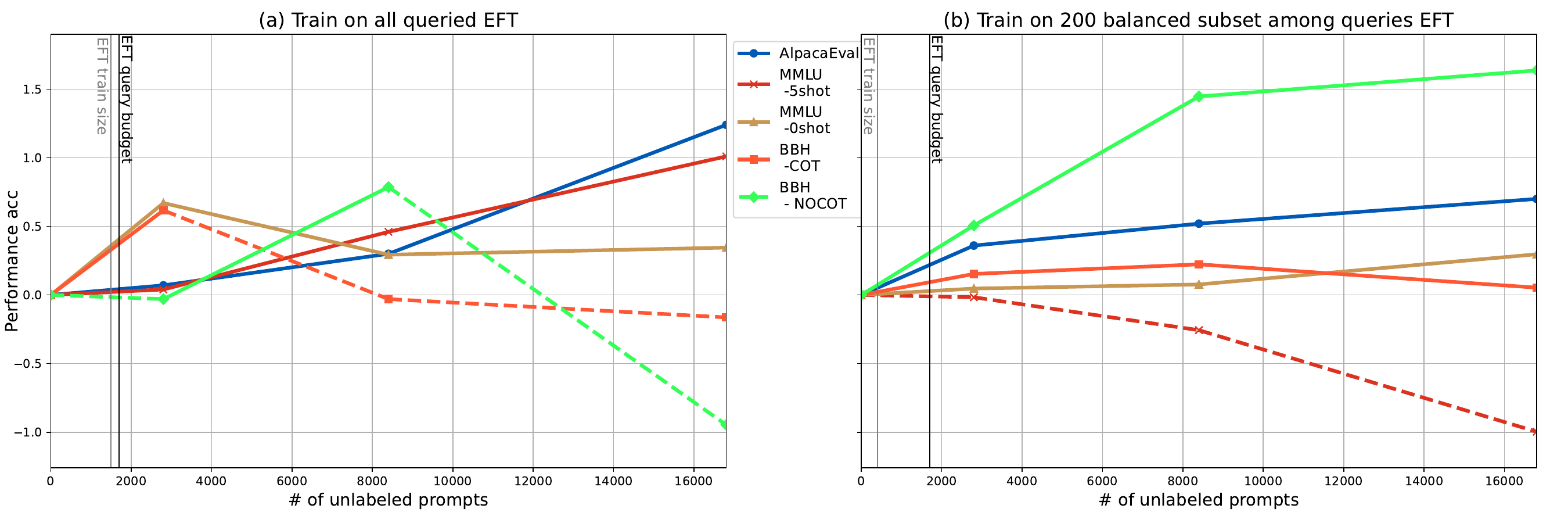}
    \caption{\textbf{With fixed query budget $n$, performance of $\meval$ across different numbers of unlabeled prompts $N$}. \textbf{Left:} The initial 1700 responses of $\bm{X}$ generated by $M_1$ are evaluated by GPT (indicated by the black vertical line). We use 1500 of these EFT data to train weak evaluators (shown in gray) and reserve the remainder as validation data to select the optimal weak evaluator. The graph displays the performance of models trained on preference sets labeled by this weak evaluator across five metrics. \textbf{Right:} Similar to the left, but instead of using the entire 1500 EFT data to train the weak evaluator, we select a balanced subset of 200 EFT as previously described.}
    \label{fig:unlabeledprompt_balance}
\end{figure*}

\paragraph{$M_\text{eval}$ trained on 1.5K $\text{EFT}_1$ can effectively label more than 9x DPO pairs.}
Despite training the evaluator $M_\text{eval}$ on EFT generated from just an initial 1.5K prompts, the evaluator is capable of effectively labeling more than nine times the number of DPO pairs, as shown in Fig.~\ref{fig:unlabeledprompt_balance} (a). This labeling capacity is mainly reflected in steady performance improvements on AlpacaEval2 and MMLU5shot. There is also a slight but less consistent improvement observed on MMLU0shot. However, the current strategies show a negative impact on BBH metrics as the number of unseen prompts increases, although an improvement of over 0.5\% is still achieved when $\text{EFT}_1$ and $\text{IFT}_1$ have greater overlap. This suggests that improvement of BBH metrics mainly comes from the ground truth feedback of the expert while the improvement of others comes from $M^\text{eval}$'s labeling ability.

\paragraph{Random on-policy with balanced training has different behaviors.} In addition, we also show the result of balanced training where we enforce the samples of each reward class to be the same during training in Fig.~\ref{fig:unlabeledprompt_balance} (b). We observe that the performance across all metrics becomes more stable under the balanced training, either showing a consistent increase or decrease. However, the magnitude of these changes is relatively modest, except for BBH-NOCOT. Interestingly, this strategy exhibits behavior opposite to that of the unbalanced version in metrics like BBH-NOCOT and MMLU5shot. The underlying reasons for these differences are not immediately clear and requires further investigation in future works. 

\begin{table}[!ht]
\setlength{\belowcaptionskip}{-20pt} 
\scalebox{0.58}{
\begin{tabular}{lccc}
\toprule
 & AlpacaEval2 & MMLU5shot & MMLU0shot \\
\midrule
\textbf{Off-policy query} & \cellcolor{orange!60} +0.84 ($\sqrt{tv}$ 0.5) & +0.02 ($\sqrt{tv}$ 0.2) & \cellcolor{orange!80} +0.66 ($\sqrt{tv}$ 0.2) \\
\midrule
\textbf{ours} & \cellcolor{orange!90}+1.33($\sqrt{tv}$0.56) & \cellcolor{orange!100}+1.26($\sqrt{tv}$0.55) & \cellcolor{orange!60}+0.54($\sqrt{tv}$0.53) \\
\bottomrule
\end{tabular}
}
\caption{\textbf{With fixed $N$, performance change from $M_1$ to $\widetilde{M}_2$ of off-policy query.} Here we choose \textbf{coresetEFT} at n=1700 as a comparison. But as explained in the setting, off-policy query comes from 3K $\EFTseed$ and therefore it should be comparable with all three methods at n=400, 1200 and 1700 in Tab.~\ref{tab:coreset}. Notably, this comparison only works under current dataset setting. In the other benign cases when $\EFTseed$ is not high biased or its rewards in known in advance, the effective number of data used in training will be close to the number of query, which suggests less queries are required to build the same size of train-set as in our non-benign case. So off-policy might gain more advantages.}
\label{tab: off-policy query}
\end{table}

\subsection{Ablation study: Compare with potential approaches adapted from previous works}
\label{sec: ablation more comparison with previous}

In this section, we further answer \textbf{Q3}. We first explore the effectiveness of our proposed methods compared to training \(\widetilde{M}^\text{eval}\) with \(\text{EFT}_\text{seed}\) (off-policy queried) and then compare to SPIN.

\begin{figure}[!h]
    \setlength{\abovecaptionskip}{0pt} 
    \setlength{\belowcaptionskip}{-20pt} 
    \centering
    \includegraphics[scale=0.07]{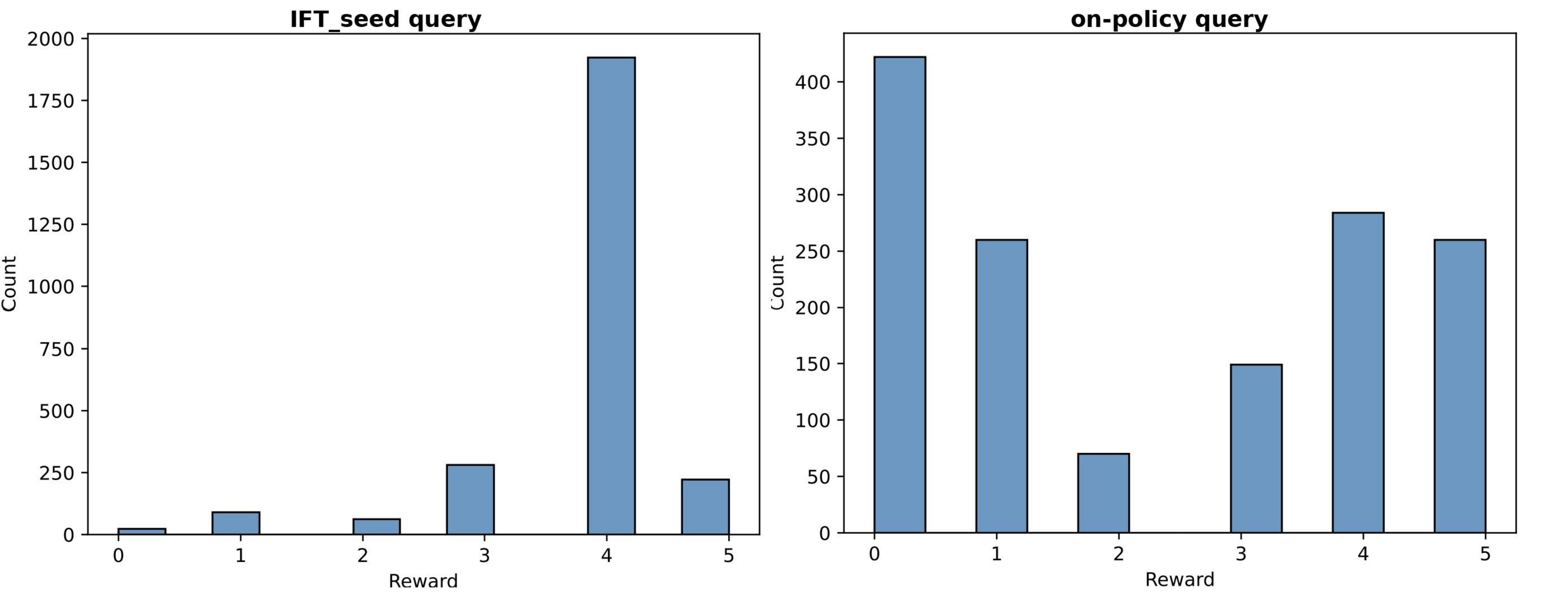}
    \caption{\textbf{Training reward distribution for $\text{EFT}_\text{seed}$ versus $\text{EFT}_1$ in our experiment,} highlighting the bias towards higher rewards in $\text{EFT}_\text{seed}$.}
    \label{fig:reward_distr}
\end{figure}

\paragraph{Balanced Nature of $\text{EFT}_1$ Compared to $\text{EFT}_\text{seed}$} One advantage of using on-policy generated samples $\text{EFT}_1$, is their naturally diverse reward distribution compared to initial $\text{EFT}_\text{seed}$, as shown in Figure~\ref{fig:reward_distr}. Consequently, we can utilize the entire queried $\text{EFT}_1$ as a training set without the need for further filtering.

\paragraph{On-policy+AL generally outperforms off-policy query} In Table~\ref{tab: off-policy query}, we show that \textbf{random on-policy} gains slight advantages compared to \textbf{off-policy}. Furthermore, adding \textbf{coresetEFT} with $n=1200$ and \textbf{coresetIFT} with $n=1700$ gains more advantages.

\paragraph{With Known Seed Rewards, Training on \(\text{EFT}_\text{seed}\) Shows Limited Advantages}
When the seed rewards are known, we can avoid wasting query costs on samples within the majority reward class, therefore only requires 500 query budget. Such \textbf{balanced off-policy query} can outperform our proposed methods when the query budget is constrained to 400 under the AlpacaEval and MMLU metrics. However, if the seed rewards distribution is unknown, the \textbf{off-policy query} method is strictly worse than our method. As the query budget increases, our proposed strategies begin to show better performance across all three metrics.

On the other hand, this advantage does not exist under more challenging metrics. For example, under BBH metrics, which are less compatible with our dataset, we show in Tab.~\ref{tab: seedIFT bad on BBH-NOCOT} that random on-policy querying still prevents a significant performance drop in the seed model \(M_1\). In contrast, training \(M^\text{eval}\) on \(\text{IFT}_\text{seed}\) results in a notable decrease in performance on BBH-NOCOT.

\begin{table}[h]
\begin{center}
\setlength{\belowcaptionskip}{-20pt} 
\scalebox{0.64}{
\setlength{\tabcolsep}{4pt} 
\begin{tabular}{lp{2.5cm}p{2.4cm}p{2.4cm}} 
\toprule
 & 200 balanced off-policy query & 200 random on-policy query & 1500 random on-policy query \\ 
\midrule
BBH-COT & +0.17 & -0.17 & +0.62 \\ 
BBH-NOCOT & \cellcolor{blue!30} -5.71 & +1.27 & +0.53 \\ 
\bottomrule
\end{tabular}
}
\caption{\textbf{Comparison between training $\text{EFT}_\text{seed}$ and our strategy under BBH metrics.}}
\label{tab: seedIFT bad on BBH-NOCOT}
\end{center}
\end{table}
\paragraph{SPIN is strictly worse in our setting.} We dmonstrate in Tab.~\ref{tab: performance comparison with SPIN} the limitations of using SPIN-related strategies with only a small initial set of labeled data can leads to negative gains. One might attribute this inferior performance of SPIN to the lack of using unlabeled prompts \(\bm{X}\). However, by comparing SPIN and SPIN+$\widetilde{M}^\text{eval}$ with $\widetilde{M}^\text{eval}$, we further discovered that these disadvantages persist even comparing with other methods under the setting that no unlabeled prompts are available. (e.g. Performance averaged over four metrics is more then 6\% worse than non SPIN one)  Therefore, we believe that performance degrades regardless of how the negative sample is chosen, as long as we fixing the ground truth $\text{IFT}_\text{seed}$ as positive sample.

\vspace{-6px}
\section{Conclusion}
This work is the first to explore cost-effective proxy reward oracle construction strategies for labeling a larger set of preferences with extremely limited labeled seed data. We identify two key techniques: on-policy query and active learning. The results convey a clear message: with a limited query budget, reward feedback should be used to train a proxy reward/preference oracle instead of being directly used for labeling.
\newpage
\section{Limitations}

\paragraph{Focus on methodology instead of achieving state-of-the-art}
This paper focuses more on the methodology rather than achieving state-of-the-art results. Therefore, we do not use more recent models like Llama3-8B or Mistral-7B, or more recent datasets like Ultra-feedback. Given that the performance of self-improvement methods highly relies on the capability of the initial pretrained model and high-quality data, our choice may limit large numerical increases. Additionally, as mentioned in Sec.~\ref{sec: strategy}, we focus only on one iteration, while most existing works validate that multiple iterations can further improve model performance. Therefore, the main direction for future work is to apply our methods in more advanced setups to achieve state-of-the-art models.

\paragraph{Limited downstream metrics}
As we mentioned earlier, the effectiveness of our algorithm highly depends on the quality of initial pretrained models and datasets. Here, we did not test on all the standard metrics like MT-bench or GSM8k since our choice of model and dataset are naturally not good at those benchmarks. After switching to more advanced setups, we should conduct a more thorough investigation.

\paragraph{Failure of using external resources}
Many existing works employ externally trained models, especially some existing reward models. It is important to combine our methods with these external resources.

\paragraph{Combine with existing iterative DPO methods}
As mentioned in Sec.~\ref{sec: existing works and baselines} and App.~\ref{app: more realted works}, many existing works assume a fixed reward/preference oracle and focus on optimizing the algorithm by proposing new loss functions or adding an extra exploratory policy. These directions seem orthogonal to our methods. It is important to combine our approach with these to see whether our approaches are truly universally applicable to all those methods.

\bibliography{anthology,custom}

\begin{thebibliography}{34}
\expandafter\ifx\csname natexlab\endcsname\relax\def\natexlab#1{#1}\fi

\bibitem[{Ash et~al.(2021)Ash, Goel, Krishnamurthy, and Kakade}]{ash2021gone}
Jordan Ash, Surbhi Goel, Akshay Krishnamurthy, and Sham Kakade. 2021.
\newblock Gone fishing: Neural active learning with fisher embeddings.
\newblock \emph{Advances in Neural Information Processing Systems}, 34:8927--8939.

\bibitem[{Ash et~al.(2019)Ash, Zhang, Krishnamurthy, Langford, and Agarwal}]{ash2019deep}
Jordan~T Ash, Chicheng Zhang, Akshay Krishnamurthy, John Langford, and Alekh Agarwal. 2019.
\newblock Deep batch active learning by diverse, uncertain gradient lower bounds.
\newblock \emph{arXiv preprint arXiv:1906.03671}.

\bibitem[{Bai et~al.(2022)Bai, Jones, Ndousse, Askell, Chen, DasSarma, Drain, Fort, Ganguli, Henighan et~al.}]{bai2022training}
Yuntao Bai, Andy Jones, Kamal Ndousse, Amanda Askell, Anna Chen, Nova DasSarma, Dawn Drain, Stanislav Fort, Deep Ganguli, Tom Henighan, et~al. 2022.
\newblock Training a helpful and harmless assistant with reinforcement learning from human feedback.
\newblock \emph{arXiv preprint arXiv:2204.05862}.

\bibitem[{Bhatt et~al.(2024)Bhatt, Chen, Das, Zhang, Truong, Mussmann, Zhu, Bilmes, Du, Jamieson et~al.}]{bhatt2024experimental}
Gantavya Bhatt, Yifang Chen, Arnav~M Das, Jifan Zhang, Sang~T Truong, Stephen Mussmann, Yinglun Zhu, Jeffrey Bilmes, Simon~S Du, Kevin Jamieson, et~al. 2024.
\newblock An experimental design framework for label-efficient supervised finetuning of large language models.
\newblock \emph{arXiv preprint arXiv:2401.06692}.

\bibitem[{Chen et~al.(2024{\natexlab{a}})Chen, Liu, Du, Pang, Liu, Sinha, Varakantham, and Lin}]{chen2024bootstrapping}
Changyu Chen, Zichen Liu, Chao Du, Tianyu Pang, Qian Liu, Arunesh Sinha, Pradeep Varakantham, and Min Lin. 2024{\natexlab{a}}.
\newblock Bootstrapping language models with dpo implicit rewards.
\newblock \emph{arXiv preprint arXiv:2406.09760}.

\bibitem[{Chen et~al.(2024{\natexlab{b}})Chen, Deng, Yuan, Ji, and Gu}]{chen2024self}
Zixiang Chen, Yihe Deng, Huizhuo Yuan, Kaixuan Ji, and Quanquan Gu. 2024{\natexlab{b}}.
\newblock Self-play fine-tuning converts weak language models to strong language models.
\newblock \emph{arXiv preprint arXiv:2401.01335}.

\bibitem[{Citovsky et~al.(2021)Citovsky, DeSalvo, Gentile, Karydas, Rajagopalan, Rostamizadeh, and Kumar}]{citovsky2021batch}
Gui Citovsky, Giulia DeSalvo, Claudio Gentile, Lazaros Karydas, Anand Rajagopalan, Afshin Rostamizadeh, and Sanjiv Kumar. 2021.
\newblock Batch active learning at scale.
\newblock \emph{Advances in Neural Information Processing Systems}, 34:11933--11944.

\bibitem[{Cook et~al.(1998)Cook, Cunningham, Pulleyblank, and Schrijver}]{cook1998combinatorial}
William~J Cook, William~H Cunningham, William~R Pulleyblank, and Alexander Schrijver. 1998.
\newblock Combinatorial optimisation.
\newblock \emph{Wiley-Interscience Series in Discrete Mathematics and Optimization, USA}, 1:998.

\bibitem[{Cui et~al.(2023)Cui, Yuan, Ding, Yao, Zhu, Ni, Xie, Liu, and Sun}]{cui2023ultrafeedback}
Ganqu Cui, Lifan Yuan, Ning Ding, Guanming Yao, Wei Zhu, Yuan Ni, Guotong Xie, Zhiyuan Liu, and Maosong Sun. 2023.
\newblock Ultrafeedback: Boosting language models with high-quality feedback.
\newblock \emph{arXiv preprint arXiv:2310.01377}.

\bibitem[{Dong et~al.(2024)Dong, Xiong, Pang, Wang, Zhao, Zhou, Jiang, Sahoo, Xiong, and Zhang}]{dong2024rlhf}
Hanze Dong, Wei Xiong, Bo~Pang, Haoxiang Wang, Han Zhao, Yingbo Zhou, Nan Jiang, Doyen Sahoo, Caiming Xiong, and Tong Zhang. 2024.
\newblock Rlhf workflow: From reward modeling to online rlhf.
\newblock \emph{arXiv preprint arXiv:2405.07863}.

\bibitem[{Gao et~al.(2024)Gao, Chang, Zhan, Oertell, Swamy, Brantley, Joachims, Bagnell, Lee, and Sun}]{gao2024rebel}
Zhaolin Gao, Jonathan~D. Chang, Wenhao Zhan, Owen Oertell, Gokul Swamy, Kianté Brantley, Thorsten Joachims, J.~Andrew Bagnell, Jason~D. Lee, and Wen Sun. 2024.
\newblock \href {http://arxiv.org/abs/2404.16767} {Rebel: Reinforcement learning via regressing relative rewards}.

\bibitem[{Geifman and El-Yaniv(2017)}]{geifman2017deep}
Yonatan Geifman and Ran El-Yaniv. 2017.
\newblock Deep active learning over the long tail.
\newblock \emph{arXiv preprint arXiv:1711.00941}.

\bibitem[{Hendrycks et~al.(2020)Hendrycks, Burns, Basart, Zou, Mazeika, Song, and Steinhardt}]{hendrycks2020measuring}
Dan Hendrycks, Collin Burns, Steven Basart, Andy Zou, Mantas Mazeika, Dawn Song, and Jacob Steinhardt. 2020.
\newblock Measuring massive multitask language understanding.
\newblock \emph{arXiv preprint arXiv:2009.03300}.

\bibitem[{Jiang et~al.(2023)Jiang, Ren, and Lin}]{jiang2023llm}
Dongfu Jiang, Xiang Ren, and Bill~Yuchen Lin. 2023.
\newblock Llm-blender: Ensembling large language models with pairwise ranking and generative fusion.
\newblock In \emph{Proceedings of the 61st Annual Meeting of the Association for Computational Linguistics (Volume 1: Long Papers)}, pages 14165--14178.

\bibitem[{K{\"o}pf et~al.(2024)K{\"o}pf, Kilcher, von R{\"u}tte, Anagnostidis, Tam, Stevens, Barhoum, Nguyen, Stanley, Nagyfi et~al.}]{kopf2024openassistant}
Andreas K{\"o}pf, Yannic Kilcher, Dimitri von R{\"u}tte, Sotiris Anagnostidis, Zhi~Rui Tam, Keith Stevens, Abdullah Barhoum, Duc Nguyen, Oliver Stanley, Rich{\'a}rd Nagyfi, et~al. 2024.
\newblock Openassistant conversations-democratizing large language model alignment.
\newblock \emph{Advances in Neural Information Processing Systems}, 36.

\bibitem[{Lambert et~al.(2024)Lambert, Pyatkin, Morrison, Miranda, Lin, Chandu, Dziri, Kumar, Zick, Choi et~al.}]{lambert2024rewardbench}
Nathan Lambert, Valentina Pyatkin, Jacob Morrison, LJ~Miranda, Bill~Yuchen Lin, Khyathi Chandu, Nouha Dziri, Sachin Kumar, Tom Zick, Yejin Choi, et~al. 2024.
\newblock Rewardbench: Evaluating reward models for language modeling.
\newblock \emph{arXiv preprint arXiv:2403.13787}.

\bibitem[{Li et~al.(2023)Li, Zhang, Dubois, Taori, Gulrajani, Guestrin, Liang, and Hashimoto}]{alpaca_eval}
Xuechen Li, Tianyi Zhang, Yann Dubois, Rohan Taori, Ishaan Gulrajani, Carlos Guestrin, Percy Liang, and Tatsunori~B. Hashimoto. 2023.
\newblock Alpacaeval: An automatic evaluator of instruction-following models.
\newblock \url{https://github.com/tatsu-lab/alpaca_eval}.

\bibitem[{Moskovitz et~al.(2023)Moskovitz, Singh, Strouse, Sandholm, Salakhutdinov, Dragan, and McAleer}]{moskovitz2023confronting}
Ted Moskovitz, Aaditya~K Singh, DJ~Strouse, Tuomas Sandholm, Ruslan Salakhutdinov, Anca~D Dragan, and Stephen McAleer. 2023.
\newblock Confronting reward model overoptimization with constrained rlhf.
\newblock \emph{arXiv preprint arXiv:2310.04373}.

\bibitem[{Ouyang et~al.(2022)Ouyang, Wu, Jiang, Almeida, Wainwright, Mishkin, Zhang, Agarwal, Slama, Ray et~al.}]{ouyang2022training}
Long Ouyang, Jeffrey Wu, Xu~Jiang, Diogo Almeida, Carroll Wainwright, Pamela Mishkin, Chong Zhang, Sandhini Agarwal, Katarina Slama, Alex Ray, et~al. 2022.
\newblock Training language models to follow instructions with human feedback.
\newblock \emph{Advances in neural information processing systems}, 35:27730--27744.

\bibitem[{Rafailov et~al.(2024)Rafailov, Sharma, Mitchell, Manning, Ermon, and Finn}]{rafailov2024direct}
Rafael Rafailov, Archit Sharma, Eric Mitchell, Christopher~D Manning, Stefano Ermon, and Chelsea Finn. 2024.
\newblock Direct preference optimization: Your language model is secretly a reward model.
\newblock \emph{Advances in Neural Information Processing Systems}, 36.

\bibitem[{Rosset et~al.(2024)Rosset, Cheng, Mitra, Santacroce, Awadallah, and Xie}]{rosset2024direct}
Corby Rosset, Ching-An Cheng, Arindam Mitra, Michael Santacroce, Ahmed Awadallah, and Tengyang Xie. 2024.
\newblock Direct nash optimization: Teaching language models to self-improve with general preferences.
\newblock \emph{arXiv preprint arXiv:2404.03715}.

\bibitem[{Sener and Savarese(2018)}]{k_center_coreset}
Ozan Sener and Silvio Savarese. 2018.
\newblock \href {https://openreview.net/forum?id=H1aIuk-RW} {Active learning for convolutional neural networks: A core-set approach}.
\newblock In \emph{International Conference on Learning Representations}.

\bibitem[{Suzgun et~al.(2023)Suzgun, Scales, Sch{\"a}rli, Gehrmann, Tay, Chung, Chowdhery, Le, Chi, Zhou et~al.}]{suzgun2023challenging}
Mirac Suzgun, Nathan Scales, Nathanael Sch{\"a}rli, Sebastian Gehrmann, Yi~Tay, Hyung~Won Chung, Aakanksha Chowdhery, Quoc Le, Ed~Chi, Denny Zhou, et~al. 2023.
\newblock Challenging big-bench tasks and whether chain-of-thought can solve them.
\newblock In \emph{Findings of the Association for Computational Linguistics: ACL 2023}, pages 13003--13051.

\bibitem[{Touvron et~al.(2023)Touvron, Martin, Stone, Albert, Almahairi, Babaei, Bashlykov, Batra, Bhargava, Bhosale et~al.}]{touvron2023llama}
Hugo Touvron, Louis Martin, Kevin Stone, Peter Albert, Amjad Almahairi, Yasmine Babaei, Nikolay Bashlykov, Soumya Batra, Prajjwal Bhargava, Shruti Bhosale, et~al. 2023.
\newblock Llama 2: Open foundation and fine-tuned chat models.
\newblock \emph{arXiv preprint arXiv:2307.09288}.

\bibitem[{Tran et~al.(2023)Tran, Glaze, and Hancock}]{viethoangtranduong}
Hoang Tran, Chris Glaze, and Braden Hancock. 2023.
\newblock Iterative dpo alignment.
\newblock Technical report, Snorkel AI.

\bibitem[{Wang et~al.(2023)Wang, Kordi, Mishra, Liu, Smith, Khashabi, and Hajishirzi}]{wang2023self}
Yizhong Wang, Yeganeh Kordi, Swaroop Mishra, Alisa Liu, Noah~A Smith, Daniel Khashabi, and Hannaneh Hajishirzi. 2023.
\newblock Self-instruct: Aligning language models with self-generated instructions.
\newblock In \emph{Proceedings of the 61st Annual Meeting of the Association for Computational Linguistics (Volume 1: Long Papers)}, pages 13484--13508.

\bibitem[{Wang et~al.(2022)Wang, Mishra, Alipoormolabashi, Kordi, Mirzaei, Arunkumar, Ashok, Dhanasekaran, Naik, Stap et~al.}]{wang2022super}
Yizhong Wang, Swaroop Mishra, Pegah Alipoormolabashi, Yeganeh Kordi, Amirreza Mirzaei, Anjana Arunkumar, Arjun Ashok, Arut~Selvan Dhanasekaran, Atharva Naik, David Stap, et~al. 2022.
\newblock Super-naturalinstructions: Generalization via declarative instructions on 1600+ nlp tasks.
\newblock In \emph{2022 Conference on Empirical Methods in Natural Language Processing, EMNLP 2022}.

\bibitem[{Wu et~al.(2024)Wu, Sun, Yuan, Ji, Yang, and Gu}]{wu2024SPO}
Yue Wu, Zhiqing Sun, Huizhuo Yuan, Kaixuan Ji, Yiming Yang, and Quanquan Gu. 2024.
\newblock Self-play preference optimization for language model alignment.
\newblock \emph{arXiv preprint arXiv:2405.00675}.

\bibitem[{Xie et~al.(2024)Xie, Foster, Krishnamurthy, Rosset, Awadallah, and Rakhlin}]{xie2024exploratory}
Tengyang Xie, Dylan~J. Foster, Akshay Krishnamurthy, Corby Rosset, Ahmed Awadallah, and Alexander Rakhlin. 2024.
\newblock \href {http://arxiv.org/abs/2405.21046} {Exploratory preference optimization: Harnessing implicit q*-approximation for sample-efficient rlhf}.

\bibitem[{Xiong et~al.(2023)Xiong, Dong, Ye, Wang, Zhong, Ji, Jiang, and Zhang}]{xiong2023iterative}
Wei Xiong, Hanze Dong, Chenlu Ye, Ziqi Wang, Han Zhong, Heng Ji, Nan Jiang, and Tong Zhang. 2023.
\newblock Iterative preference learning from human feedback: Bridging theory and practice for rlhf under kl-constraint.
\newblock In \emph{ICLR 2024 Workshop on Mathematical and Empirical Understanding of Foundation Models}.

\bibitem[{Xu et~al.(2023)Xu, Lee, Sukhbaatar, and Weston}]{xu2023some}
Jing Xu, Andrew Lee, Sainbayar Sukhbaatar, and Jason Weston. 2023.
\newblock Some things are more cringe than others: Preference optimization with the pairwise cringe loss.
\newblock \emph{arXiv preprint arXiv:2312.16682}.

\bibitem[{Yuan et~al.(2024)Yuan, Pang, Cho, Sukhbaatar, Xu, and Weston}]{yuan2024selfreward}
Weizhe Yuan, Richard~Yuanzhe Pang, Kyunghyun Cho, Sainbayar Sukhbaatar, Jing Xu, and Jason Weston. 2024.
\newblock Self-rewarding language models.
\newblock \emph{arXiv preprint arXiv:2401.10020}.

\bibitem[{Zheng et~al.(2024)Zheng, Chiang, Sheng, Zhuang, Wu, Zhuang, Lin, Li, Li, Xing et~al.}]{zheng2024judging}
Lianmin Zheng, Wei-Lin Chiang, Ying Sheng, Siyuan Zhuang, Zhanghao Wu, Yonghao Zhuang, Zi~Lin, Zhuohan Li, Dacheng Li, Eric Xing, et~al. 2024.
\newblock Judging llm-as-a-judge with mt-bench and chatbot arena.
\newblock \emph{Advances in Neural Information Processing Systems}, 36.

\bibitem[{Zhu et~al.(2023)Zhu, Frick, Wu, Zhu, and Jiao}]{zhu2023starling}
Banghua Zhu, Evan Frick, Tianhao Wu, Hanlin Zhu, and Jiantao Jiao. 2023.
\newblock Starling-7b: Improving llm helpfulness \& harmlessness with rlaif.

\end{thebibliography}
\bibliographystyle{acl_natbib}

\newpage
\onecolumn
\appendix
\section{More related works}
\label{app: more realted works}

\subsection{Efficient query with fixed reward/preference oracle} Many existing works focus on efficient query by assuming the reward/preference oracle is good enough. In the other word, they want to select the data that is informative for training the policy model, which is the generative tasks, instead of informative for training the evaluation model, which is the discriminative tasks. This motivation is highly related classical reinforcement learning topics. Specifically, works such as \cite{xu2023some, touvron2023llama} start using iterative DPO with the same loss as original DPO paper. Later works like \citet{rosset2024direct, wu2024SPO, gao2024rebel, xie2024exploratory} proposes to use more advanced loss instead of DPO loss. (\cite{rosset2024direct} propose new loss in their theoretical section but in their experiment they still use something like original DPO loss). \citet{chen2024bootstrapping} also employs a self-improvement style algorithm; however, instead of relying on a general reward, they construct a reward that implicitly debiases based on length. Finally, while all of those above works are focusing on on-policy query, \cite{xiong2023iterative, dong2024rlhf} further propose to maintain an extra exploratory strategy to cover more space, therefore combine the on-policy and off-policy strategy.

\subsection{Pipeline for SPIN and direct query}
\begin{figure*}[!h]
    \centering
    \setlength{\belowcaptionskip}{-18pt} 
    \includegraphics[scale=0.6]{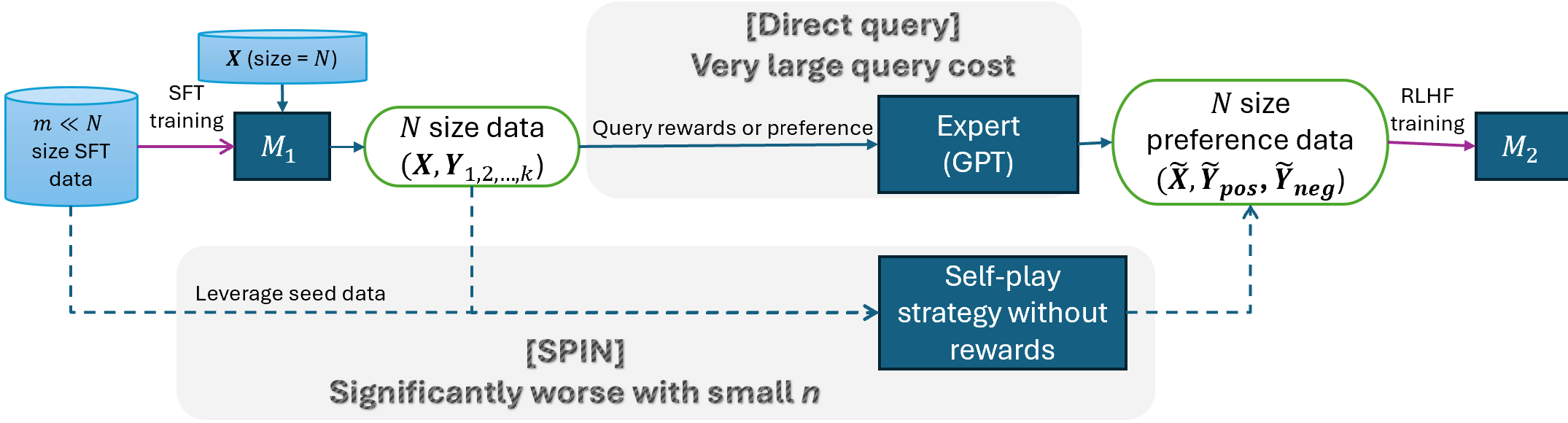}
    \caption{\textbf{Previous preference data labeling pipelines.} The figure depicts two methods, \textbf{direct query} and \textbf{SPIN}, both of which do not require proxy reward oracles. And thus the direct query demand high budget while SPIN is strictly outperforms by our methods when $m$ is small. }
    \label{fig: pipeline_previous}
\end{figure*}

\section{More experimental setup}
\label{app: experiment setup}

\subsection{Hyperparameters}
\paragraph{Hyper-parameter for training $M_1$, $M_2$} When training $M_1$, we use SFT training pipelines with batch size 128, 2 epochs, learning rate $2e-5$ and warmup rate $0.03$. When training $M_2$, we use DPO training pipelines with batch size 32, 2 epochs for 16800 number of unlabeled prompts and 3 epoch for others, learning rate $5e-7$ and warmup rate $0.01$. 

\paragraph{Hyper-parameter for training $M^\text{eval}$ and $\widetilde{M}^\text{eval}$} When training the evaluation models using either on-policy generated $\text{EFT}_1$ or $\text{EFT}_\text{seed}$, we use the same setting as training $M_1$. 

\paragraph{Hyper-parameter for generating $\text{EFT}_1$ and corresponding DPO} For each instruction in $\bm{X}$ and $\text{IFT}_\text{seed}$ (when compare with SPIN), we generate $k=4$ responses with $\text{maxLength} = 1024$. Then to give the reward feedback for each generated response, we call $M^\text{eval}$ three times (therefore get at most three EFT) and compute the average reward. We do not explicitly use the justification feedback, but such justification serves as chain-of-thought to help generate proper reward. Also not all response can received rewards, sometimes the $M^\text{eval}$ can fail to give any reasonable evaluation. In that case, we will directly discard the sample. Therefore, among 16.8K prompts, we only get about 15K DPO pairs.

\subsection{Example of EFT}
\label{app: example of EFT}
We use the exact same approach as present in Figure.2 in \cite{yuan2024selfreward} and therefore omit the details here.

\section{More experimental results}
\label{app more results}

\newpage
\subsection{Visualization of Table.~\ref{tab:coreset}}
\begin{figure}[!h]
    \centering
    \includegraphics[width=0.6\linewidth]{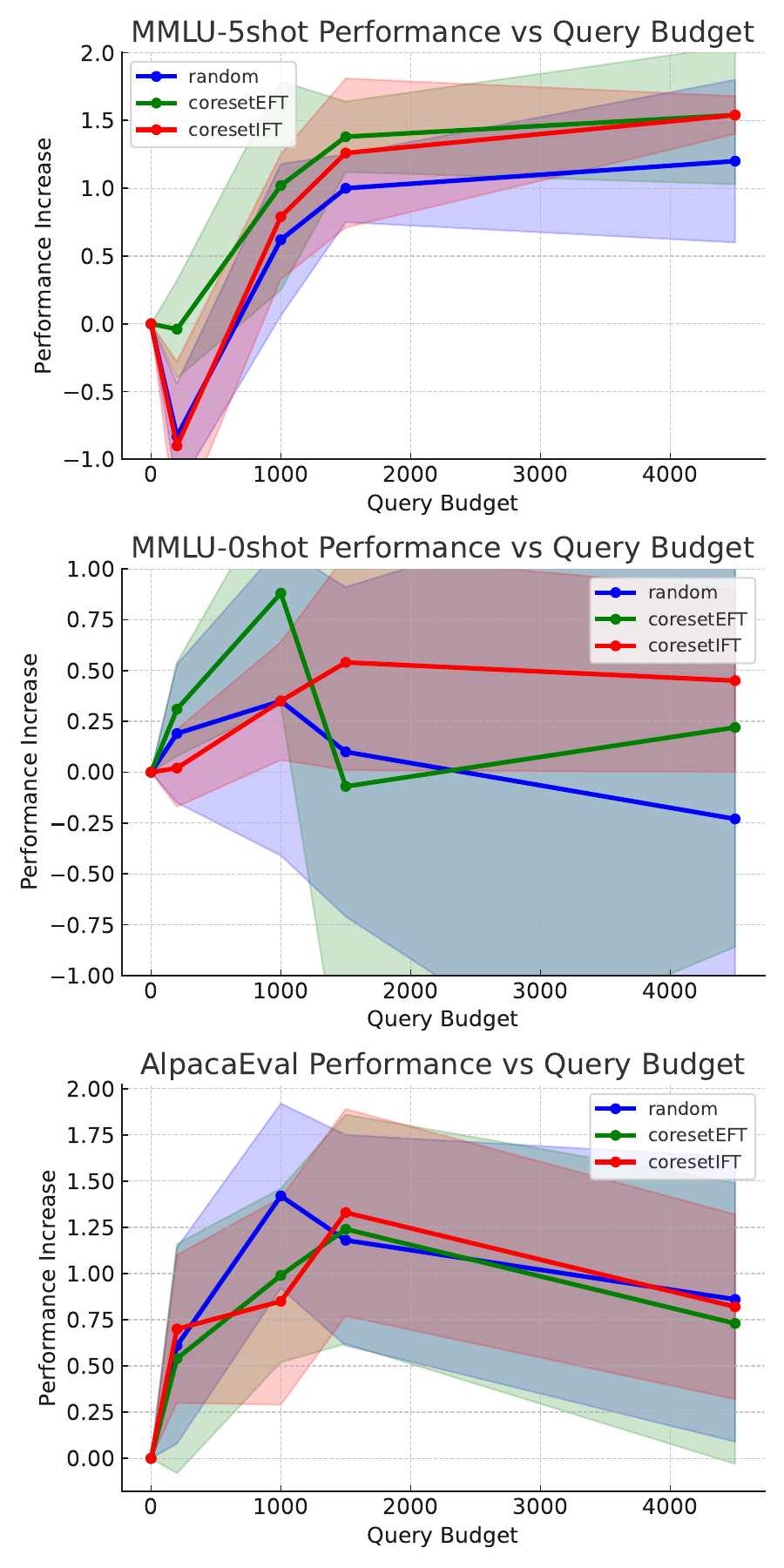}
    \captionof{figure}{\textbf{$M_2$ Performance vs Query Budget.} The shade represent the square root of total variance.}
    \label{fig: performance_vs_query_budget}
\end{figure}

we show the visualization of Table.~\ref{tab:coreset} in Fig.~\ref{fig: performance_vs_query_budget}.

\subsection{Ablation Study: Compare with previous work} 

\paragraph{Fixed Ground Truth as Positive Sample Significantly Reduces Performance}
SPIN is strictly worse than our proposed methods as shown in Tab.~\ref{tab: performance comparison with SPIN}. One might attribute this inferior performance of SPIN to the lack of using unlabeled prompts \(\bm{X}\), we argue that the core issue persists even when using only \(\text{EFT}_\text{seed}\) and \(\text{IFT}_\text{seed}\). In Tab.~\ref{tab: performance comparison with SPIN}, by comparing SPIN and SPIN+$\widetilde{M}^\text{eval}$ with $\widetilde{M}^\text{eval}$, we clearly see that, even with the same seed prompts, the performance degrades regardless of how the negative sample is chosen, as long as we fixing the ground truth $\text{IFT}_\text{seed}$ as positive sample. We hypothesize that this decline is caused by the very limited budget of initial seed data, which likely leads to over-fitting.


\begin{table}[!h]
\begin{center}
\setlength{\belowcaptionskip}{-20pt} 
\scalebox{0.7}{
\setlength{\tabcolsep}{4pt} 
\begin{tabular}{lp{2cm}p{2cm}p{2.3cm}} 
\toprule
& SPIN& SPIN+$\widetilde{M}^\text{eval}$&$\widetilde{M}^\text{eval}$   \\
\midrule
MMLU-0shot & \cellcolor{blue!20}-0.61 & \cellcolor{blue!20}-0.58 & +0.06 \\
MMLU-5shot & \cellcolor{blue!20}-1.72 & \cellcolor{blue!20}-1.85 & -0.13 \\
BBH-COT & \cellcolor{blue!20} -1.39 & \cellcolor{blue!20}-1.38 & +0.29 \\
BBH-NOCOT & \cellcolor{blue!40} -14.9 & \cellcolor{blue!40} -17.69 & -1.58 \\
\bottomrule
\end{tabular}
}
\caption{\textbf{Comparison of performance under different strategies when only using the prompt from $\text{IFT}_\text{seed}$}.}
\label{tab: performance comparison with SPIN}
\end{center}
\end{table}

\subsection{Ablation Study: Correlation between validation loss $\meval$ and held-out metrics}
\label{sec: ablation study}

We further explore the correlation between the negative validation loss of \(M^\text{eval}\) (neg\_EvalLoss) and downstream performance metrics.

\begin{figure}[h]
    \centering
    \begin{minipage}[b]{0.35\textwidth}
        \centering
        \includegraphics[width=\textwidth]{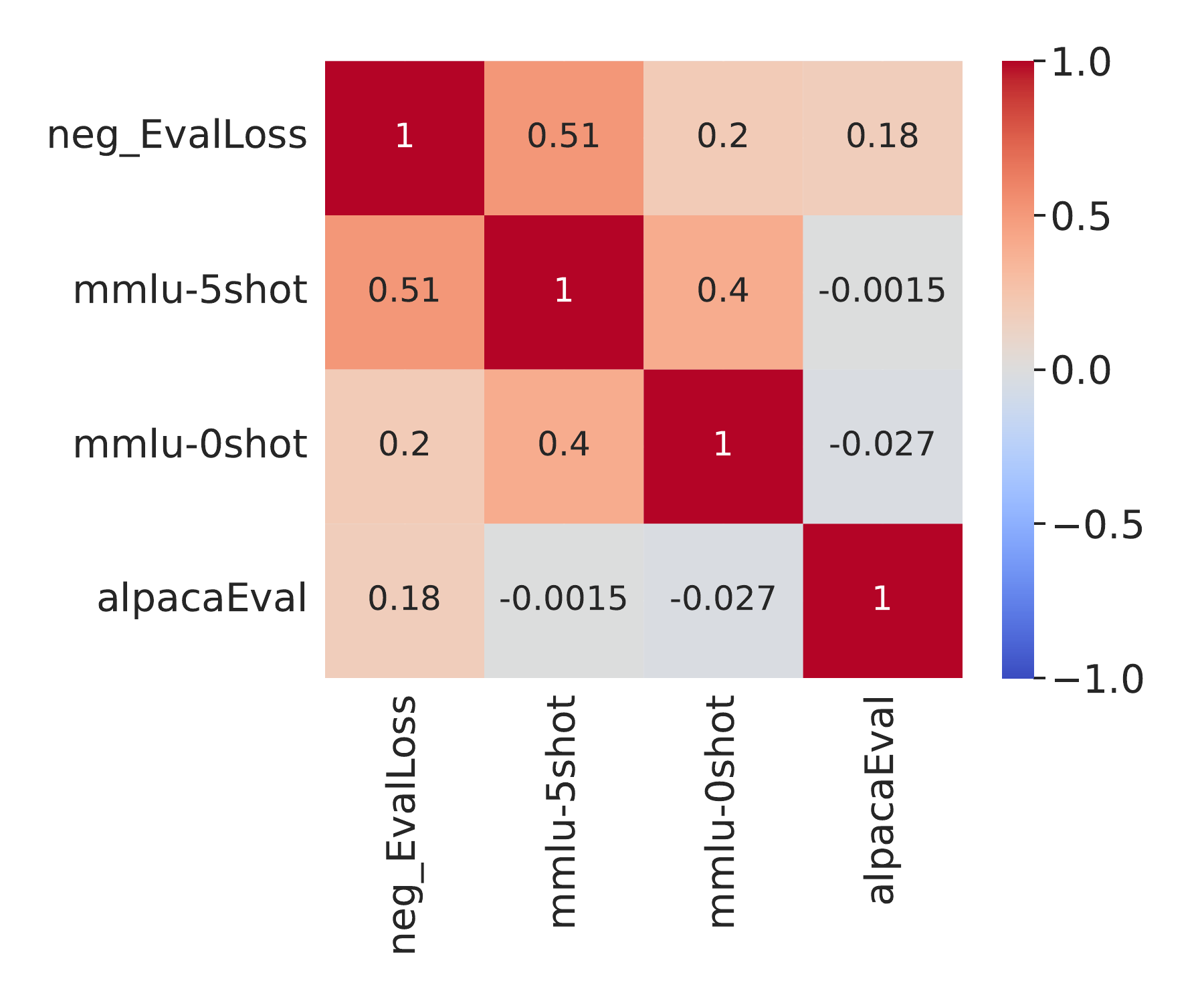}
        \label{fig:correlation_matrix}
    \end{minipage}
    \begin{minipage}[b]{0.4\textwidth}
        \centering
        \includegraphics[width=\textwidth]{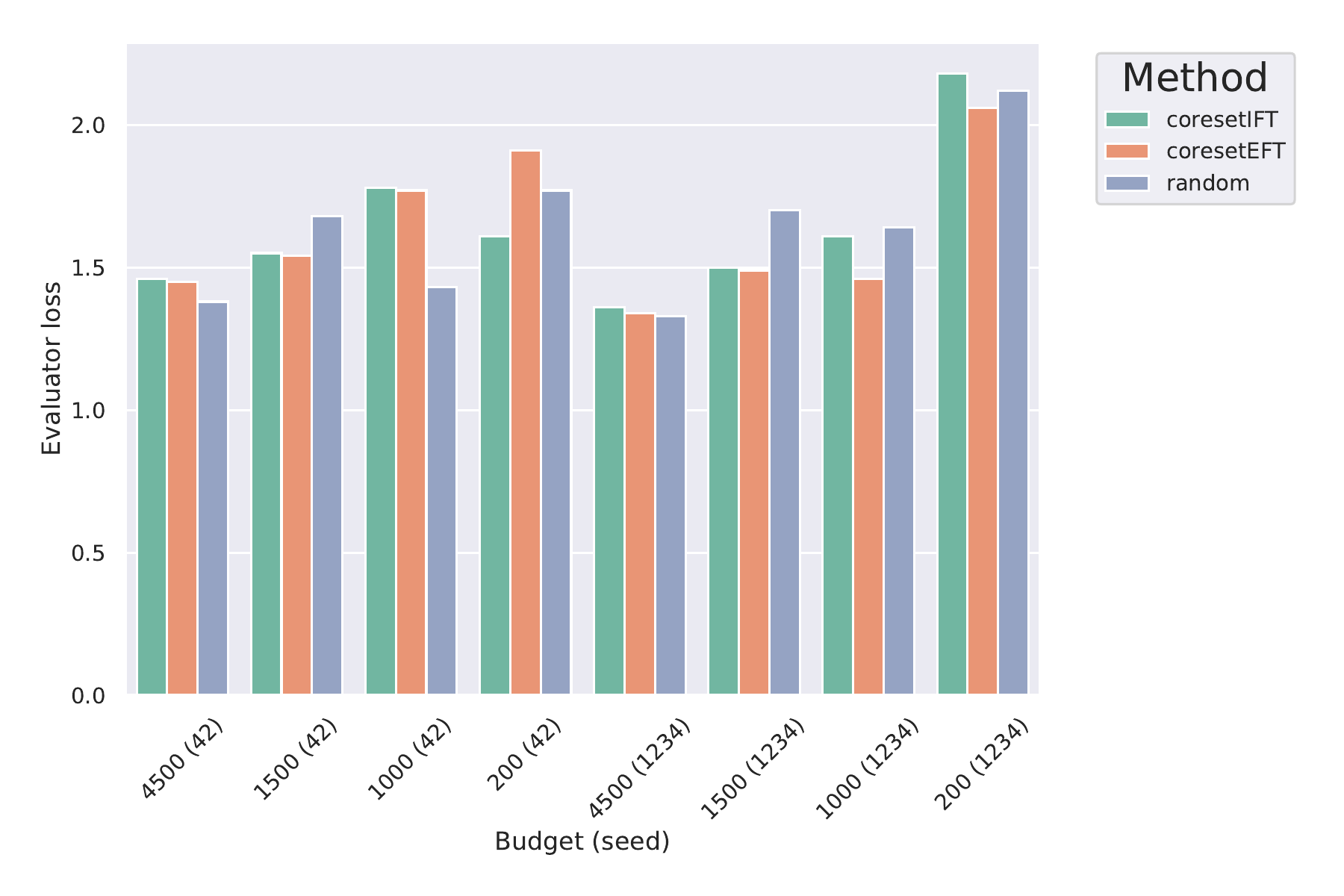}
        \label{fig:eval_loss_vs_budget}
    \end{minipage}
    \caption{\textbf{Top:} \textbf{Correlation matrix between each metric and the negative validation loss of $M^\text{eval}$.} \textbf{Bottom:} \textbf{Validation loss of of $M^\text{eval}$ across different query budgets, \(M_1\) training seeds, and strategies.} For representational purposes, we show results for two seeds, 42 and 1234.}
    \label{fig:evalLoss vs metrics}
\end{figure}

\paragraph{Metrics with Stronger Correlation to \(M^\text{eval}\) Performance Benefit most from AL Strategies}
As illustrated in top pf Fig.~\ref{fig:evalLoss vs metrics}, MMLU-5shot, which demonstrates the most stable performance and the most significant benefits from AL strategies, exhibits the strongest correlation with the validation loss of \(M^\text{eval}\). In contrast, the other two metrics, which exhibit signs of over-optimization, have much weaker correlations.

\paragraph{AL Strategies Do Not Necessarily Lead to Lower Validation Loss}
Given the above observations, one might naturally assume that AL strategies would result in a lower validation loss for \(M^\text{eval}\), thereby leading to better outcomes compared to random on-policy strategies. However, contrary to expectations, our results as depicted in bottom of Fig.~\ref{fig:evalLoss vs metrics}, show that the validation loss appears quite random. This suggests that the advantages of active querying may not stem directly from an overall improvements in \(M^\text{eval}\) performance but rather from more nuanced factors.

\subsection{Ablation study: Influence from the initial $M_1$}
\label{app: influnce form the inital $M_1$}
\begin{table}[!h]
\begin{center}
\scalebox{0.8}{
\begin{tabular}{lccc}
\toprule
 & random & coresetEFT & coresetIFT \\ 
\multicolumn{4}{c}{query budget = 200} \\ 
\midrule
AlpacaEval2 & 0.53 & 0.63 & 0.2 \\ 
MMLU-5shot & 0.18 & 0.12 & 0.19 \\ 
MMLU-0shot & 0.1 & 0.16 & 0.04 \\ 
\midrule
\multicolumn{4}{c}{query budget = 1000} \\ 
\midrule
AlpacaEval2 & 0.46 & 0.48 & 0.34 \\ 
MMLU-5shot & 0.08 & 0.14 & 0.05 \\ 
MMLU-0shot & 0.08 & 0.11 & 0.03 \\ 
\midrule
\multicolumn{4}{c}{query budget = 1500} \\ 
\midrule
AlpacaEval2 & 0.3 & 0.55 & 0.55 \\ 
MMLU-5shot & 0.11 & 0.15 & 0.07 \\ 
MMLU-0shot & 0.18 & 0.19 & 0.11 \\ 
\midrule
\multicolumn{4}{c}{query budget = 4500} \\ 
\midrule
AlpacaEval2 & 0.51 & 0.49 & 0.5 \\ 
MMLU-5shot & 0.11 & 0.08 & 0.13 \\ 
MMLU-0shot & 0.1 & 0.15 & 0.08 \\ 
\bottomrule
\end{tabular}
}
\caption{\textbf{Standard deviation averaged over different $M_1$.} Consistent with the settings in Table~\ref{tab:coreset}, this measure computes the average standard deviation of $M_2$ across different corresponding $M_1$ models, reflecting the variability in DPO training.}
\label{tab: coreset std}
\end{center}
\end{table}

The total variance in some metrics, especially MMLU0shot, is considerable. By combining data from Tab.~\ref{tab:coreset} and Tab.~\ref{tab: coreset std}, we show that for the AlpacaEval metric, both the randomness in the initial \(M_1\) training and \(M_2\) training contribute to the final variance. However, for MMLU0shot and MMLU5shot, the variance mainly stems from the randomness in the initial \(M_1\) training. Section~\ref{app: influnce form the inital $M_1$} provides a further investigation into how variability in \(M_1\) performance affects overall outcomes.

\begin{figure}[!h]
    \centering
    \includegraphics[scale=0.6]{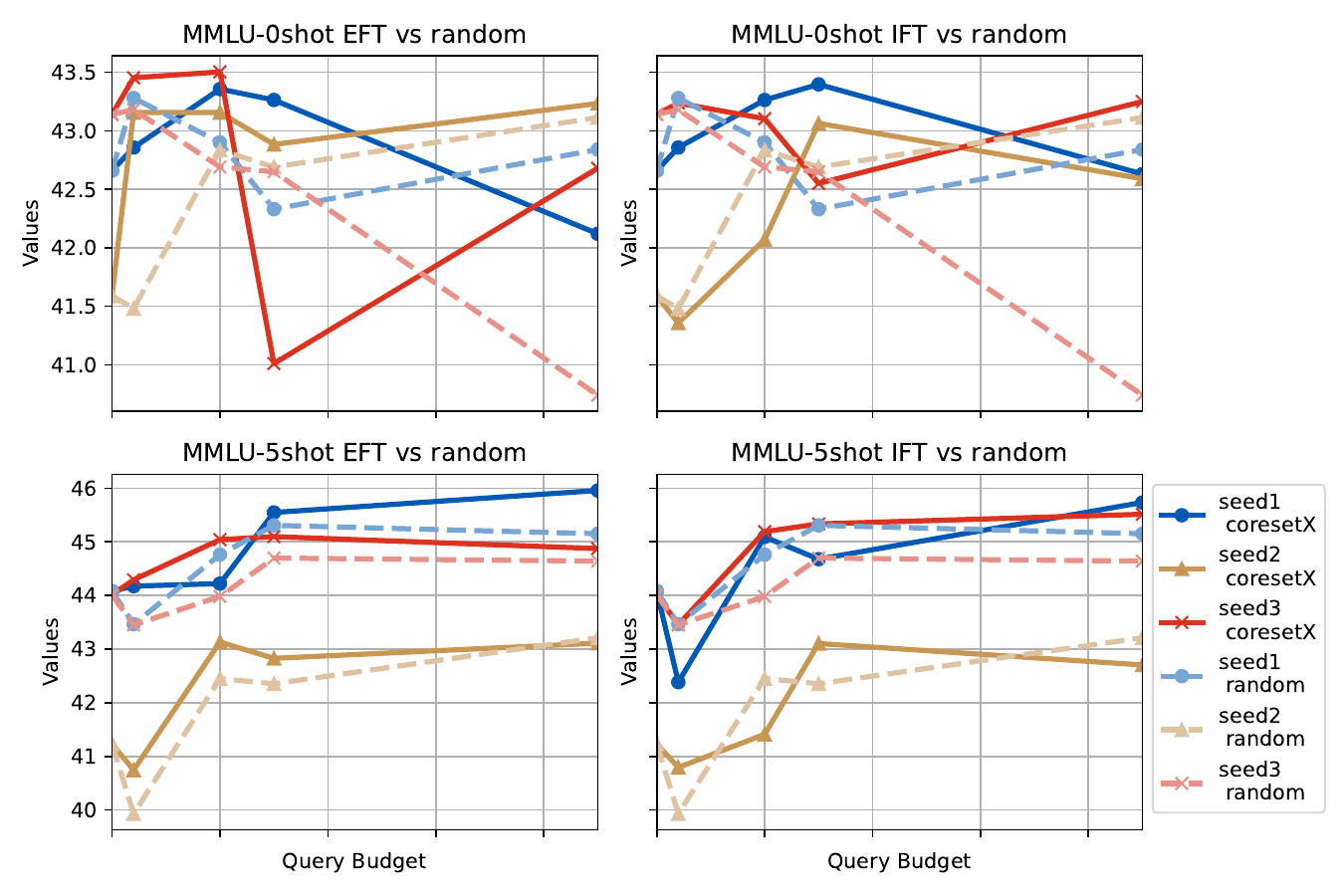}
    \caption{\textbf{With fixed N, performance change on conditioned on different $M_1$ under MMLU metrics.} Here we show the performance change by using coresetEFT(Left) and coresetIFT(Right) under three different $M_1$, whose initial performances varies. The setting is the same as in Fig.~\ref{fig: performance_vs_query_budget}. The dash lines represent all random strategies, which is the same on left and right part of the figure; while the solid line represent the coresetEFT strategy on the left and coresetIFT strategy on the right. }
    \label{fig: coreset seperate}
\end{figure}

Here we show the performance curve separated under different initial seeds in Fig.~\ref{fig: coreset seperate}.

\paragraph{A good $M_1$ is important for the effectiveness of active learning in MMLU-5shot}
After comparing the active strategy with the random strategy conditioned on different $M_1$, we observe that the advantage of active querying only occurs when the initial $M_1$ performance is strong. When initial $M_1$ has bad performance as shown in seed 2, the active strategy will become close or even worse then random.

\paragraph{The general trends between EFT and IFT are similar}
Although there are fluctuations of different magnitudes, the general trends between the two embeddings are similar. For example, in MMLU-0shot, both the seed3 active learning strategy and seed1 show a decrease in the middle followed by a rise, while seed1 increases in the middle and then downgrades.

\paragraph{For all three methods, the largest improvement occurs when the performance of $M_1$ is initially poor}
For all three on-policy methods, regardless of whether the active learning strategy is used, the largest improvement occurs when the initial $M_1$'s performance is hindered by the random seed. This is expected because the self-improvement does not introduce any new responses as new knowledge. Instead, it tends to boost the intrinsic capacity of the model itself.

\end{document}